# Wearing face mask detection using deep learning through COVID-19 pandemic

**Javad Khoramdel[a], Soheila Hatami[b], and Majid Sadedel[c,*]**
a. *Faculty of Mechanical Engineering*, *Tarbiat Modares University, Tehran, Iran.* (E-mail address: j.khoramdel@modares.ac.ir)
b. *Faculty of Mechanical Engineering*, *Tarbiat Modares University, Tehran, Iran.* (E-mail address: soheila.hatami@modares.ac.ir)
c. *Faculty of Mechanical Engineering*, *Tarbiat Modares University, Tehran, Iran.* (Tel.: +98(21)82884987, Fax: +98(21)82884987, E-mail address: majid.sadedel@modares.ac.ir) – Corresponding author.

**Abstract**. During the COVID-19 pandemic, wearing a face mask has been known to be an effective way to prevent the spread of COVID-19. In lots of monitoring tasks, humans have been replaced with computers thanks to the outstanding performance of the deep learning models. Monitoring the wearing of a face mask is another task that can be done by deep learning models with acceptable accuracy. The main challenge of this task is the limited amount of data because of the quarantine. In this paper, we did an investigation on the capability of three state-of-the-art object detection neural networks on face mask detection for real-time applications. As mentioned, here are three models used, Single Shot Detector (SSD), two versions of You Only Look Once (YOLO) i.e., YOLOv4-tiny, and YOLOv4-tiny-3l from which the best was selected. In the proposed method, according to the performance of different models, the best model that can be suitable for use in real-world and mobile device applications in comparison to other recent studies was the YOLOv4-tiny model, with 85.31% and 50.66 for mean Average Precision (mAP) and Frames Per Second (FPS), respectively. These acceptable values were achieved using two datasets with only 1531 images in three separate classes.
**KEYWORDS:** COVID-19, Deep Learning, Object Detection, Face Mask, Convolutional Neural Networks

## 1. Introduction

It has been more than two years since the COVID-19 pandemic (known as the coronavirus) has been started and the disease is still at its peak in some regions of the world. With the continuing of this pandemic, the situation is deteriorating in many countries. According to the World Health Organization (WHO) [1], people should wear masks in all crowded and even secluded environments that are in contact with others. Many people follow this law well, but there are still so many individuals who do not follow this law and risk their lives and the lives of others. For this reason, many governments deal with these violators after identifying them as face mask detection in smart cities [2]. One way to identify these people is to use manual methods to monitor and control them, but unfortunately this is not possible, especially for crowded environments, and it is a tedious task. Accordingly, in this paper, we have gone to an automated method, which is the use of artificial intelligence and deep learning. The use of object-based models in deep learning allows us to distinguish people who wear a mask correctly, people who do not wear a mask, and people who wear a mask incorrectly, each of whom can use a variety of masks also. This method can even be used to obtain compliance with health protocols for any place like hospitals.

The remainder of this work is as follows. In the next section we briefly review related works toward face mask detection. Details of the used dataset and the proposed algorithm are given in section 3, where we describe the two datasets that we merged them and also the models and networks that we used during our work. The final results and evaluations are presented in section 4. Finally, section 5 summarizes our conclusions.

The main contributions of the proposed method are as follows:

- Due to the lack of a dataset compromising people with hijab and similar face to Iranian people, the proposed method has been able to achieve a suitable mean Average Precision (mAP) and Frames Per Second (FPS) using two prepared datasets with less similarity and even with a small number of images.
- Also, in this method, there is no need for preprocessing during train and test phases, and use three classes instead of two unlike other previous works to simulate real-word situations more.
- Finally, by comparing three different models based on Convolutional Neural Network (CNN), the best of which is with a low computational cost and volume space for storage to obtain a suitable model that is appropriate for mobile device applications.

2## 2. Related works

To distinguish people with mask, many parts of the face are covered. For this reason, it is not possible to easily recognize the face and then go to diagnose the way people wear mask (with mask, without mask, and incorrect mask). In fact, in [3], people's faces are identified using Multi-task Cascaded Convolutional Networks, and in the output, five landmarks are marked on the face. Recognizing these five landmarks means fully recognizing people's faces. The method that the authors of this article have used to recognize the face is practical in some ways, but unfortunately, it cannot be easily used to identify the faces with masks because many of these landmarks are covered. For example, in [4], the authors prepared their dataset, which includes people with correct masks and incorrect masks by using landmarks. This dataset contains almost the same distribution of both classes. Also, they have synthetically created this dataset. In this way, they have used the method described in [5] for the basis of the work and put a mask on people's faces. They used 68 landmarks on a face without a mask and 12 landmarks on the mask alone. If these 12 landmarks are placed on the landmarks that fit the face, it means that the person has worn the mask correctly, otherwise, the mask should have covered part of the face and this means that the mask has not been worn correctly.

Rodriguez et al. [6] have proposed a system that triggers an alarm if personnel in the operating room do not wear mandatory masks. The system consists of two face and mask detectors that use the tone in HSV color space. The proposed system has reached a recall above 95% with a false positive below 5% for the detection of faces and surgical masks.

In [7], the authors used a model which consists of two parts. For feature extraction, they used ResNet-50 and for classification, they used three different classifiers which are decision trees, Support Vector Machine (SVM), and ensemble algorithm. Also, they worked on three different datasets which are publicly available and they obtained good results over these datasets which were described in detail, but they did not report any value for FPS to know about its real-time application.

In [8], four popular object detection algorithms, two of them from You Only Look Once (YOLO) family i.e., YOLOv3, YOLOv3-tiny, Single Shot Multibox Detector (SSD), and Faster Region-Based Convolutional Neural Network (Faster R-CNN) were used to recognize people with and without mask, and for this work, they used their own dataset, Moxa3k [9] which consists of 3000 images from Kaggle dataset of medical mask dataset [10] and others were collected from the internet. They concluded that YOLOv3-tiny is more suitable for a good balance of accuracy and real-time application, which they need a way to have a reasonable accuracy for real-time work with CCTV cameras especially to detect people in crowded places. Loey et al. [11] utilized the InceptionV3 pre-trained model by removing its last layer and adding five layers for fine-tuning the model. They also used simulated Face Mask Dataset (SMFD) [12] which has a good balance in two classes (simulated masked facial and unmasked facial images) for their work and they achieved reasonable results for both test and train phases.

In some other works like in [13], authors used some prepared libraries such as TensorFlow, Keras, and OpenCV which are simple enough but are not deep to extract features for complicated conditions and make them unsuitable for some real-world situations.

In [14], the authors have used different models including Faster R-CNN, YOLOv3, YOLOv4, YOLOv5, and YOLOR, for face mask detection and also social distance determination. Besides that, they have collected a dataset consisting of several different video frames with two classes. The ViDMASK includes 20,000 instances of people with mask and 2,500 instances without mask; which clearly shows the non-uniform distribution of the samples across different classes. It should be noted that although the number of the images in the ViDMASK is larger than Moxa3k dataset, they were collected from videos. Hence, there are not many variations between different frames.

## 3. The proposed method

In this section, the datasets used as well as the proposed method are described in detail.

### 3.1. Dataset

Here, for our purpose, we used two different prepared datasets. Of course, we have made changes to these two datasets to prepare them for our final results. The first dataset [15] contains 853 images in PNG, jpg, and jpeg format. This dataset has three classes of people with correct mask, without mask, and incorrect mask. It contains people in different situations including in a crowded environment where the number of people in the image is high and also where there is only one person in the image. The annotation of these images by default is all in the form of an XLM file. This XLM file includes the name of the image folder, the name of the image to which the class belongs, width, height, depth (RGB images), image class name (with mask, without mask, and incorrect mask), bounding box coordinate (includes $x_{min}$, $y_{min}$, $x_{max}$, and $y_{max}$), which the first two are the top-right coordinates of the bounding box and the next two are the bottom left coordinates of the bounding box. Therefore, for working with YOLO and SSD models, we have converted these XML files to txt files. The next dataset [10] contains 678 images in jpg format. The images in this dataset also include three classes of people with correct mask, without mask, and incorrect mask. Also, the annotations for the images in this dataset include annotations in XLM and txt files. The XLM files of these images contain the same data as the previous dataset.

**Figure 1.**

Finally, we merged these selected datasets. Our final dataset has images in jpg and PNG format. Of all the dataset images, 80% are for the train data and 20% for the test data. Also, Figure 11 contains some samples of the dataset images.
In Figure 12 the frequency of instances in each class is shown. As it can be seen, the frequency of the "incorrect mask" class instances is too low in comparison with two other classes. In the presented dataset the number of "with mask", "without mask", and "incorrect mask" are 6322, 1377, and 247, respectively.

### *3.2. Object detection*

In this paper, the goal is to develop a measure for observing the necessary protocols for COVID-19 or any place which needs to obtain compliance with health protocols. To do this, at the first stage, an object detection algorithm is employed to detect and classify people with mask, without mask and those who have worn the mask incorrectly. Lots of algorithms have been introduced in recent years for object detection and lots of progress has been made in this field. Since AlexNet won the ImageNet challenge in 2012 [16], CNNs have gained momentum in computer vision tasks. CNN-based approaches have reached to a significant superiority to the non-CNN-based algorithms [17] like HoG [18], SIFT [19], Haar feature-based object detection [20], etc. It should be noted that like many other tasks, there is a tradeoff between speed and accuracy. Some CNNs like Faster R-CNN have achieved high accuracy but suffer from the low speed at the run time [21] which makes it unsuitable for real-time applications.

**Figure 2.**

To speed up the object detection task, YOLO was developed by Redmon et al. [22] in 2016. Inspired by YOLO, SSD was arisen in 2016 [23]. However, SSD provided higher accuracy and speed in comparison with YOLO, so far, four different versions of YOLO have been released and each version was modified to obtain higher accuracy and run time speed. Each version of YOLO also has a lighter version which is called YOLO-tiny. The YOLO-tiny network uses fewer convolutional layers in comparison with YOLO which results in an increase in speed by sacrificing the accuracy. However, when the dataset doesn't have enough samples for training, YOLO-tiny may have a better performance than YOLO on that dataset. After the success of SSD to outperform YOLO by doing object detection on the multiple scales of the feature map to handle the problem with small size objects, YOLOv3 [24] was presented with this idea to perform object detection on multiple scales to obtain higher accuracy. YOLOv3 uses three scales for this purpose and 9 anchor boxes per each grid cell while YOLOv3-tiny uses only two scales and six anchor boxes per each grid cell. Adarsh et al. [25] claimed that while the mAP of YOLOv3-tiny is 33.2% on COCO dataset [26], the mAP of YOLOv3 is 57.8% on it; but YOLOv3-tiny can perform object detection 11 times faster than YOLOv3.
By the time YOLOv4 was proposed by Bochkovskiy et al. [27], two structures for YOLOv4-tiny were introduced. Same as YOLOv3-tiny, one of them uses two scales for object detection and six anchor boxes per each grid cell (this version is called YOLOv4-tiny) but the other version uses three scales for object detection and nine bounding boxes per each grid cell (this version is called YOLOv4-tiny-3l, since it uses three scales for detection). The YOLOv4-tiny-3l is expected to be slower than YOLOv4-tiny and faster than YOLOv4. Since there are no comparisons between the accuracy and speed of SSD, YOLOv4-tiny, and YOLOv4-tiny-3l, these networks will be trained on the above-mentioned dataset to choose the efficient object detector for this dataset in terms of speed and accuracy. After doing object detection on a single frame, instances of each class are counted and the distance between the detected people in the image space is calculated.

## 4. Experimental results

The training process has been conducted on a device with one Tesla K80 GPU and a single-core hyper-threaded Xeon Processor 2.3 GHz. The resolution of the input image was 416x416 for both YOLOv4-tiny networks and 300x300 for the SSD. For the networks to converge faster and have a more generalization power, the pre-trained weights on COCO dataset were used as the initial weights for the training. The dataset for mask detection, contained 1540 images, 1232 of them were used for training and 308 images for validation. After the training, seven images were collected from the internet as the test images. The result of the prediction for the YOLOv4-tiny-3l, YOLOv4-tiny, and SSD on these images are shown respectively in Figure 13, Figure 14 and Figure 15. These test images contain both crowded and uncrowded. Some of these images were taken outdoors (a, b, e, and g) and indoors (c, d, and f) under different lighting conditions. As it can be seen in Figure 13 c, there was a person with incorrect mask in the image but the YOLOv4-tiny-3l has failed to identify it and classified that person as a person "without mask". However, Figure 14 c shows that the YOLOv4-tiny has done a better job in this case and classified that person correctly. In Figure 15, the bounding box for the classes "with mask", "without mask" and "incorrect mask" are shown with the colors green, red and orange, respectively. As it can be seen, the SSD model did not detect any people with the "incorrect mask" class, exactly the same class that has the least number of images in the dataset. While the dataset mainly includes the images from east Asian individuals (Figure 11) and did not contain pictures of Iranian people, YOLOv4-tiny and YOLOv4-tiny-3l succeeded in face mask detection on Iranian people (Figure 13-c, d, and e, Figure 14-c, d, and e). As a consequence, trained models are applicable for surveillance in Iran. Whether the images are crowded or not, YOLOv4-tiny and YOLOv4-tiny-3l detected the desired classes correctly unless the faces were blurred or very far from the camera.

Table 9 shows the Average Precision (AP) of the networks on the validation data. These results have been calculated with an IOU threshold of 50%. As it can be seen, although the YOLOv4-tiny-3l slightly outperforms the YOLOv4-tiny on the "with mask" and "without mask" classes, the YOLOv4-tiny has a significantly better performance on the "incorrect mask" class which has led to a better mAP. The YOLOv4-tiny has achieved a desirable mAP which indicates that the object detection task on this dataset is applicable. On the other hand, the lower mAP of two other models (YOLOv4-tiny-3l and SSD) which are heavier in comparison with the YOLOv4-tiny and were expected to have higher accuracy shows that the size of the dataset was not suitable for training these heavier models.

Table 10 shows the speed of these three models on the test phase. As was expected, the YOLOv4-tiny model which has the least parameters in comparison with the two other models has achieved the highest FPS. Besides, the FPS of all these three models is good enough for real-time applications. Besides, since many surveillance cameras have a frame rate of 30, the FPS of all three models is good enough for real-time applications. It should be noted that the average human walking speed is between 1.2 and 1.4 $m/s$. Hence, it is unnecessary to process all the 30 frames taken by the camera, and only considering one-third of the frames would be enough (10 frames).

**Figure 3.**

As it can be seen in Figure 12, the "incorrect mask" class instances are much less than the instances of the "with mask" and "without mask". One may say that maybe there is no difference between "incorrect mask" and "without mask" classes in terms of the health and safety protocols. However, the visual characteristics of a person with an incorrect mask are more similar to a person with mask rather than a person without mask. To validate this assumption, we merged the "incorrect mask" and "without mask" classes in one class, then we trained the YOLOv4-tiny on "without mask" and "with mask" classes. The results of evaluating the model on the validation data have been shown in

Table 11. As it can be seen, the mAP has decreased in comparison with the previous part. So, this was not a good idea to merge "incorrect mask" and "without mask" class. This result also implies another benefit of the trained models in this paper over the works considering two classes. Not only was the "incorrect mask" class detected, but also the accuracy of the model on two other categories increased. In other words, the trained models with three classes are more robust.

**Figure 4.**

On the same dataset, Roy et al. [8] trained YOLOv3, YOLOv3-tiny, SSD, and Faster R-CNN [8]. The highest reported mAP for the Intersection Over Union (IOU) threshold of 50% is 66.84% for YOLOv3 with an input image resolution of 608x608. Although the YOLOv3 model is a more complex model than the YOLOv4-tiny and also YOLOv4-tiny-3l models, our trained models have shown a better performance. As it was mentioned earlier, the resolution for input images was 416x416 which reduces the computational cost even more in comparison with the resolution of 608x608 and improves the speed at the test phase. Hence, our model is more accurate and faster than the trained model in [8].

Although Loey et al. [11] claimed they are doing face mask detection, their methodology did not involve object detection. In fact, they are doing the face detection in the first phase, then after finding the faces in the image, they are cropped and fed into an InceptionV3 model for the image classification task. They did not mention which object detection they are using for the first phase. If they are using CNNs for face detection in the first stage, then this is not efficient at all to run two heavy models (one for face detection, the other one for face mask classification) in cascade mode and will slow down the prediction, especially in crowded images. Our results show that it is possible for an object detection network to detect faces with mask and without mask accurately. If they used the classic algorithms for face detection like Haar cascade, then not only the face detection accuracy drop-offs especially in crowded areas, but also because of the nature of these algorithms, they can't take advantage of GPU for parallel computations, so the overall speed and accuracy will be highly affected. Finally, our proposed algorithm is superior to what Loey et al. proposed in both cases.

**Figure 5.**

**Table 1.**

**Table 2.**

Also, the proposed method in [3] is applicable only on small images (they used the resolution of 12x12 for the input images). This means their approach doesn't have the required scalability to be able to detect faces that are far from the camera and can be only used on the close-up faces in the image. This scalability issue also occurs in the work done by Rodriguez et al. [6].

**Table 3.**

As Table 12 shows, the proposed method has been compared with a number of recent studies. The method used in each of these recent studies is all presented with the advantages and disadvantages and finally their mAP and FPS. In the methods section, each of the studies has used different networks and architectures, and studies have shown that all the work that has been done in this field has not changed the architectures or even the networks, except in some parts, parameters were just changed. In fact, it is clear that in any recent work, as well as in the proposed method, a comparison has been made between the best networks and architectures.

One of the fundamental issues in each of these methods is the use of datasets with a large number and variety of images, except for one case of all studies, other cases have used prepared datasets, and this is a great advantage for them. While, for example, there is no data for people with hijab and with the variety of head and face covering that exists in Iran, and for this reason we were forced to use two publicly available datasets, even though they do not resemble people with hijab. Finally, some datasets used in these works have used images in which there are images of real people's faces and have artificially used masks on their faces so they prepared or used synthetic data even for test phase.

In some cases, although the method used in [7] and [28] was discussed as real-time, no value was reported for FPS. For this reason, this claim is almost unacceptable. Although in [14], high accuracy of 92.4% was reported on ViDMASK with YOLOR model, the highest accuracy on Moxa3k was obtained with Faster R-CNN model with the accuracy of 74.7%. After Faster R-CNN which is one of the slowest models (16 FPS), the most accurate model was YOLOv4-tiny (68.22%) that is considerably faster than Faster R-CNN (139 FPS). It should be noted that these accuracies were reported on detecting two classes, while the more accurate results were obtained in our study for detecting three and two classes.

Other issues, such as the fact that some methods require data preprocessing, are another drawback of previous studies. This is done before training phase or even in some cases for the test phase. In some methods, networks and architectures with a high computational cost have been used, which of course has been mentioned in some previous methods. Given that the issue of face mask detection is a topic that requires an algorithm with light computational cost (for example, for use in mobile device applications), so our proposed method presented an acceptable model that can give the best results with fewer parameters. Finally, it can be noted that one of the things that is less mentioned in the previous methods is the issue of the number of classes with a variety that can be according to the real-world. The use of three classes is another advantage of the proposed method, which can be used not only in the case of COVID-19 pandemic, but also in places such as hospitals, personnel who wear the mask incorrectly, are detected and this helps the health protocols as much as possible.

**Table 4.**

## 5. Conclusions

The main goal of this paper was an investigation on the performance of three recent state-of-the-art object detection neural networks on the face mask detection task. It was shown that although there are a limited amount of data for face mask detection, it is possible to detect and classify people in the images in three classes: "with mask", "without mask" and "incorrect mask". These networks showed acceptable performance on the test images even in crowded areas. It was also shown that it is possible to use these networks for object detection in real-time scenarios.

Finally, it can be concluded that for mobile device applications (with low space for storage) and, of course, for real-world images or video frames, even with a small number and variety of data, the proposed method has been able to strike a balance between the mAP and FPS.

One of the ways that can help improve this method in the future is to use datasets with a higher variety of images and their similarity to the Iranian people (if this method wants to be defined, for example, for people with a special type of face). This method can also be used for places where they can be identified even with a mask on their face. For example, to identify people with mask or another cover on their face in order to enter the workplace. In the end, it is possible to benefit from datasets such as ViDMASK [14] and by balancing its data and using the optimized models obtained in this article, better results for other environments could be achieved.

Figures' captions:

**Figure 6.** Some samples of the dataset images

**Figure 7.** Frequency of each class in the face mask detection dataset

**Figure 8.** The performance of YOLOv4-tiny-3l on test images

**Figure 9.** The performance of YOLOv4-tiny on test images

**Figure 10.** The performance of SSD on test images

Tables' captions:

**Table 5.** Accuracy of the models on the validation data

**Table 6.** Accuracy of the models on the validation data

**Table 7.** Accuracy of the models on the validation data

**Table 8.** Comparison of some recent studies with the proposed method

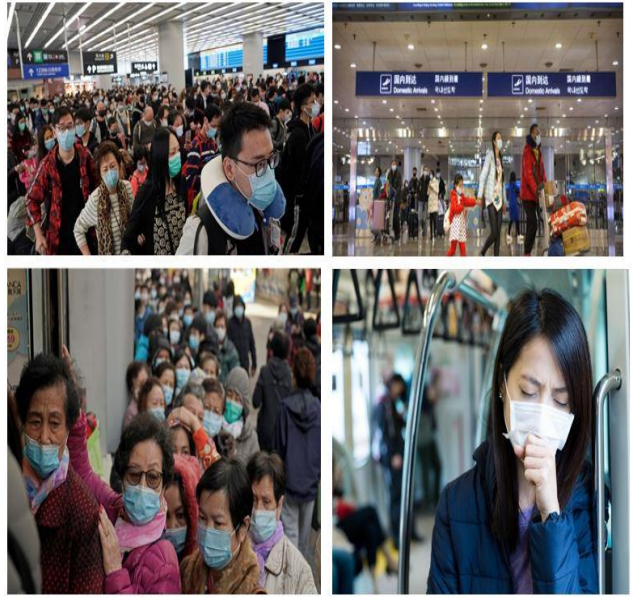

**Figure 11.**

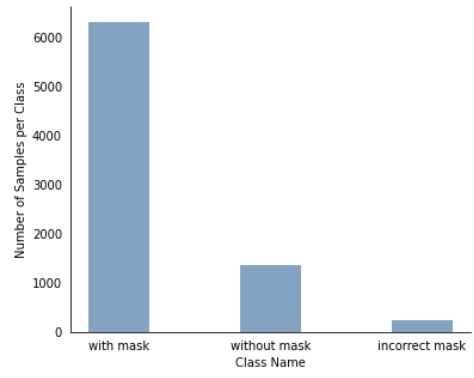

**Figure 12.**

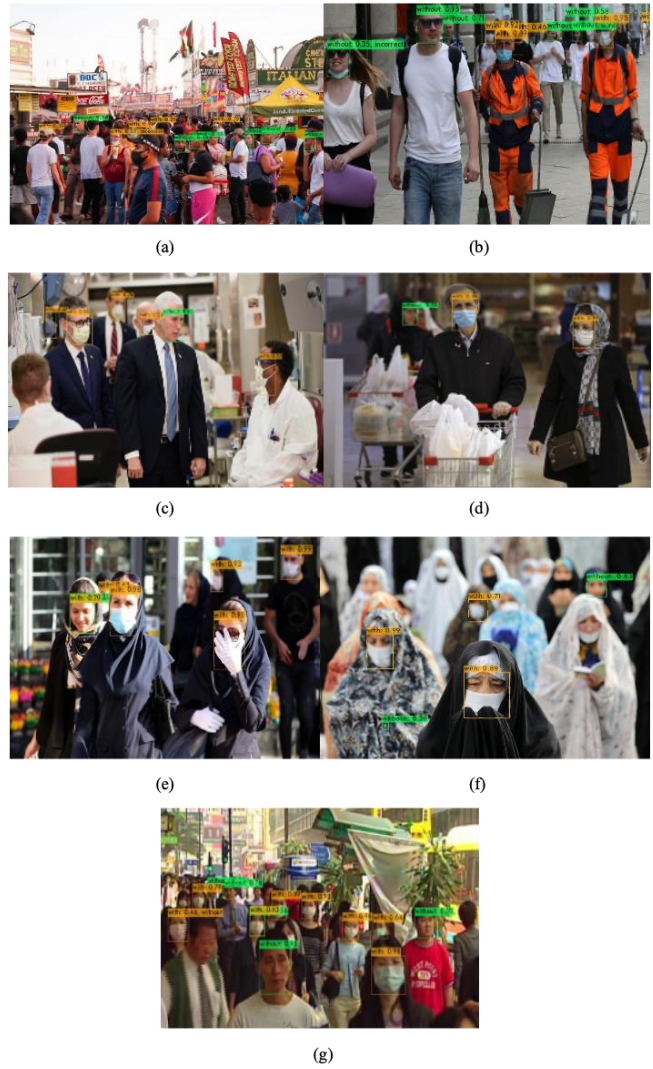

**Figure 13.**

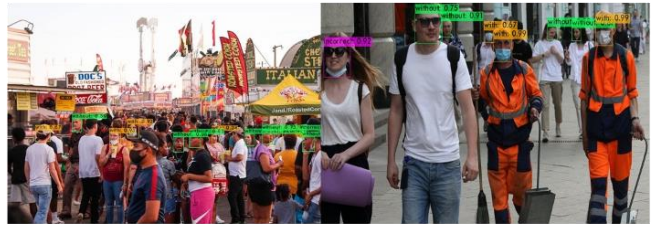

(a) (b)

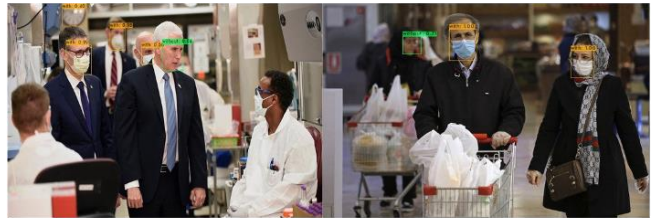

(c) (d)

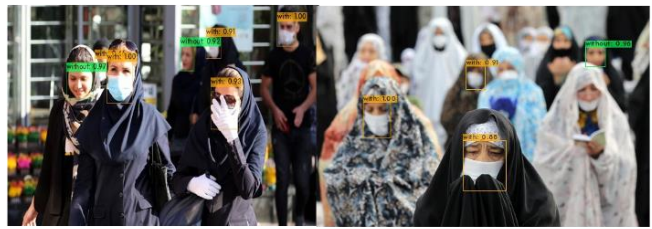

(e) (f)

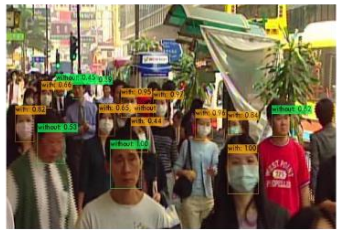

(g)

**Figure 14.**

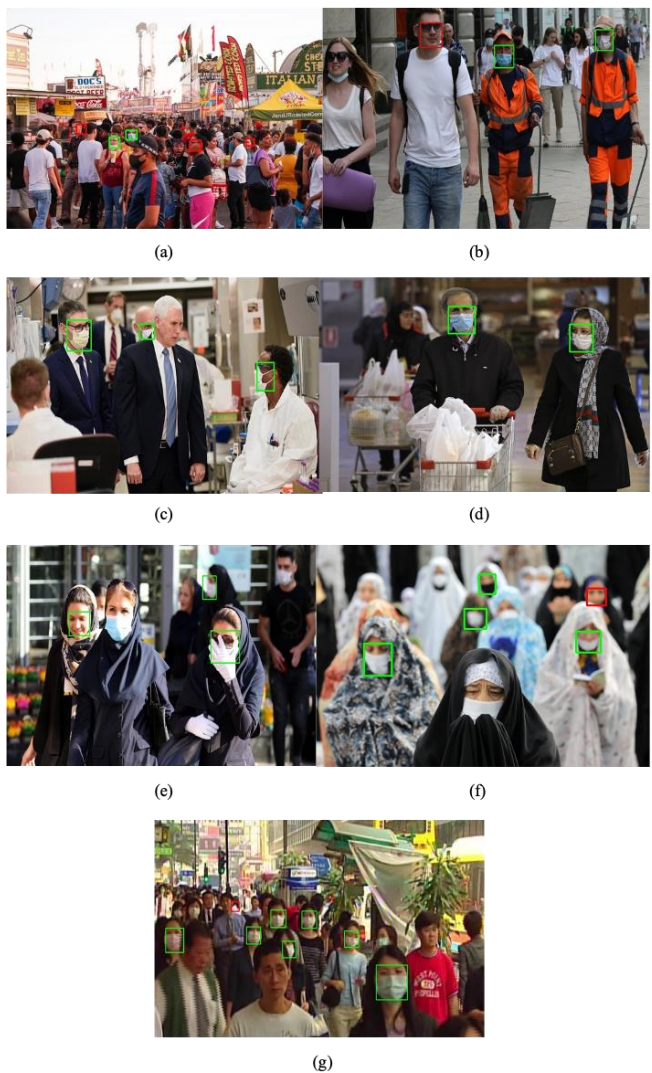

**Figure 15.**

**Table 9.**

|  | AP | | |
| --- | --- | --- | --- |
| Class | YOLOv4-tiny-3l | YOLOv4-tiny | SSD |
| Incorrect mask | 77.53% | 90.48% | 54.72% |
| With mask | 94.87% | 90.37% | 79.47% |
| Without mask | 78.62% | 75.08% | 47.20% |
| mAP | 83.67% | 85.31% | 60.46% |

**Table 10.**

| Model | YOLOv4-tiny-3l | YOLOv4-tiny | SSD |
| --- | --- | --- | --- |
| Speed (FPS) | 38.3 | 50.66 | 27.14 |

**Table 11.**

| Class | AP |
| --- | --- |
| | YOLOv4-tiny-3l |
| With mask | 89.82% |
| Without mask | 61.33% |
| mAP | 75.57% |

**Table 12.**

| Study | Method | Advantages | Disadvantages | mAP, FPS |
| --- | --- | --- | --- | --- |
| Loey et al. [7] | ResNet50 (feature extractor) + SVM, ensemble, decision tree (classifier, use them as separate classifiers) | Use three prepared different datasets with 10000, 1570 and 13000 images each | Use preprocessing, high computational cost, use a dataset with only one labeled face per image, not suitable for real-time applications, use a variety of synthetic data for testing, contains only two classes (mask and no mask) | For SVM (the best) 99.64% (real-world dataset), 99.49% (synthetic dataset), 100% (synthetic dataset), FPS not reported |
| Roy et al. [8] | SSD, F-RCNN, YOLOv3, YOLOv3-tiny | Prepare and use a dataset with 3000 images | Achieve a lower mAP, use networks with high computational cost and use some models with being not applicable for real-time, contains only two classes (mask and no mask) | 56.27% (for YOLOv3-tiny, real-world dataset), 138 |
| Asghar et al. [28] | (Preprocessing) DS-CNN (Depthwise Separable Convolutional Neural Network) | Use three prepared different datasets with 8000-combination of the first two datasets and 3000 images for the third one | Not suitable for real-time applications (the FPS is not reported), use a variety of synthetic data for testing (8000 images), contains only two classes (mask and no mask) | 92% (synthetic dataset), FPS not reported |
| Ottakh et al. [14] | Faster R-CNN, YOLOv4, YOLOv5, YOLOR | ViDMASK dataset was collected and different models were trained and evaluated on ViDMASK and Moxa3k | Only two classes were considered, low accuracy was reported on Moxa3k | 92.4% (YOLOR) on ViDMASK, 68.2% (YOLOv4 tiny) on Moxa3k, 139 (YOLOv4 tiny), 16.9 (YOLOR) |
| The Proposed Method | SSD, YOLOv4-tiny-3l, YOLOv4-tiny | No need for preprocessing, contains three classes (mask, no mask and incorrect mask), have low computational cost, contains low space for storage, suitable for mobile device applications | Dataset with a limited number of images | 85.31% (real-world dataset), 50.66 |

## Biographies

**Javad Khoramdel** is currently pursuing his MSc degree in Mechatronics at Tarbiat Modares University, Tehran, Iran. He received his BSc degree in Mechanical Engineering from K. N. Toosi University of Technology, Tehran, in 2019. His research interests are in computer vision, robotics, and deep learning.

**Soheila Hatami** received her MSc degree in Mechatronics from Tarbiat Modares University, Tehran, Iran, in 2022. She obtained her BSc degree from University of Mohaghegh Ardabili, Ardabil, and Imam Khomeini International University (as a guest student), Qazvin, in 2017, all in Electrical Engineering (power branch). Her research interests are in deep learning, image processing, industrial automation, and autonomous vehicles.

**Majid Sadedel** is currently an Assistant Professor at Tarbiat Modares University, Tehran, Iran. He received his PhD degree from Tehran University, Tehran in 2016, MSc degree from Sharif University of Technology, Tehran in 2011, and BSc degree from Amirkabir University of Technology, Tehran in 2009, all in Mechanical Engineering. His research interests are in robotics, artificial intelligence, mechatronics, and industrial automation.